\documentclass[10pt,twocolumn,letterpaper]{article}

\usepackage[pagenumbers]{cvpr} 

\usepackage{array}
\usepackage{multirow}
\usepackage{colortbl}
\usepackage{tabularx}
\usepackage{tikz}
\usetikzlibrary{arrows.meta,positioning,calc,fit}

\newcommand{\loss}{\mathcal{L}}

\usepackage{catchfile}
\makeatletter
\newcommand{\MAWAMInputBibliography}{%
  \IfFileExists{\jobname.bbl}{%
    \newcommand{\MAWAMBbl}{\input{\jobname.bbl}}%
    \patchcmd{\MAWAMBbl}
      {$$\backslash$mu\_0 $}{$\mu_0$}{}{}%
    \MAWAMBbl
  }{\@input@{\jobname.bbl}}%
}
\patchcmd{\bibliography}
  {\@input@{\jobname.bbl}}{\MAWAMInputBibliography}{}%
  {\PackageError{ma-wam}{Cannot install bibliography display repair}
    {Check the definition of \string\bibliography.}}
\makeatother

\definecolor{cvprblue}{rgb}{0.21,0.49,0.74}
\usepackage[pagebackref,breaklinks,colorlinks,allcolors=cvprblue]{hyperref}

\def\confName{CVPR}
\def\confYear{2027}

\title{Trace-WAM: Learning to Act via 3D Motion Prediction for World Action Models}
\author{
Xu Hu$^{1,2}$\quad Changjie Wu$^{2}$\quad Linzhe Shi$^{2}$\quad Hang Zhang$^{2}$\quad Chen Yang$^{2,3}$\\
Lingjun Zhang$^{2,3}$\quad Guowen Zhang$^{1}$\quad Mu Xu$^{2}$\quad Junran Peng$^{5}$\quad Qing Li$^{1,\dagger}$\quad Zhaoxiang Zhang$^{4,\dagger}$\\[5pt]
{\footnotesize $^{1}$The Hong Kong Polytechnic University\quad $^{2}$Amap, Alibaba Group\quad $^{3}$Tsinghua University}\\
{\footnotesize $^{4}$NLPR, Institute of Automation, Chinese Academy of Sciences (CASIA)\quad $^{5}$University of Science and Technology Beijing}\\[3pt]
{\small Project Page: \url{https://trace-wam.github.io/}}
}
\makeatletter
\g@addto@macro\@maketitle{
\noindent\begin{minipage}{\textwidth}
\captionsetup{type=figure}
\centering
\includegraphics[width=\textwidth]{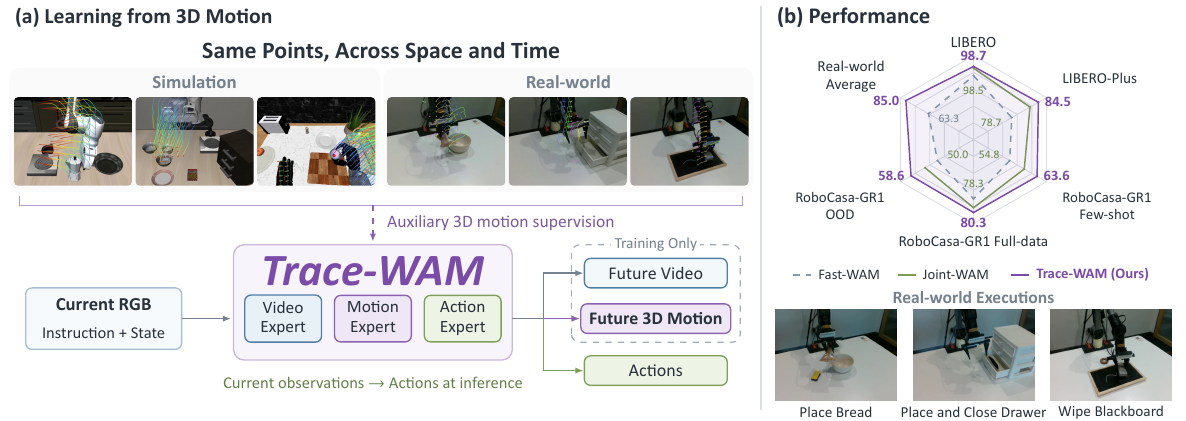}
\caption{\textbf{Overview of Trace-WAM.} (a) Motion annotations from simulation and real-world demonstrations provide auxiliary 3D motion supervision for Trace-WAM. The model jointly learns video, motion, and action prediction; future video and motion prediction are used only during training. (b) Benchmark performance and real-world policy executions.}
\label{fig:teaser}
\end{minipage}\par
\vspace{\textfloatsep}
}
\makeatother
\begin{document}
\maketitle
{\renewcommand{\thefootnote}{\fnsymbol{footnote}}\footnotetext[2]{Corresponding authors.}}
\begin{abstract}
World Action Models (WAMs) learn representations for robot control by coupling future visual prediction with action generation. Successful manipulation depends primarily on how the robot and task-relevant objects evolve, whereas future-video prediction must also reconstruct the appearance and background of the entire scene. We therefore propose Trace-WAM, which augments video-action learning with future 3D motion prediction as an auxiliary objective. This objective supervises temporal point correspondence together with image-plane displacement and depth change, directly describing the interaction dynamics and spatial changes involved in manipulation. To this end, we adopt a three-branch Mixture-of-Transformer (Tri-MoT) architecture to jointly learn video prediction, motion prediction, and action generation. The motion expert predicts future trajectories from current visual features, allowing motion supervision to shape the representations used for action generation. Because the action expert does not consume predicted trajectories, inference retains the current-observation policy path: a single visual forward pass followed by action denoising. Experiments on RoboCasa-GR1, LIBERO, and LIBERO-Plus demonstrate strong in-distribution performance and out-of-distribution generalization, with particularly clear gains over Fast-WAM under limited demonstrations and distribution shifts. 
\end{abstract}

\section{Introduction}
\label{sec:intro}

Vision-Language-Action (VLA) models~\cite{kim2024openvla,pertsch2025fast,intelligence2025pi_} adapt pretrained vision-language representations to robot control, learning to generate actions from visual observations and language instructions. Despite their strong policy-learning capabilities, most VLAs focus on action prediction without explicitly modeling how the scene evolves during execution. World Action Models (WAMs) incorporate this future evolution into policy learning by combining future visual prediction with action generation~\cite{guo2026unified,yuan2026fast}. Built on pretrained video models, WAMs transfer rich spatiotemporal priors to robot manipulation by learning from future observations together with action sequences.
Yet successful manipulation depends primarily on how the robot and task-relevant objects change over time, whereas future RGB prediction must also model texture, illumination, and background appearance across the entire scene. 
This difference motivates a complementary target that emphasizes interaction dynamics while retaining the rich visual context learned from future RGB.
Recent approaches enrich policy representations by augmenting visual prediction with auxiliary supervision for geometry, semantics, and interaction structure. These objectives include depth and geometric prediction, semantic feature learning, and supervision of affordances and goal regions~\cite{guo2026unified,ma2026geosem,yuan2026dreamwam,zhang2026deva,zheng2026gift,li2026wam4d}, encouraging representations to capture spatial layout and task-relevant interaction constraints. A per-frame or region-level target, however, does not by itself identify how the same physical locations evolve throughout an interaction, a distinction also emphasized by 4D-WAM~\cite{yang20264d}. This temporal correspondence connects the current scene to the sequence of changes required to complete a manipulation task.

\begin{figure}[t]
\centering
\includegraphics[width=\linewidth]{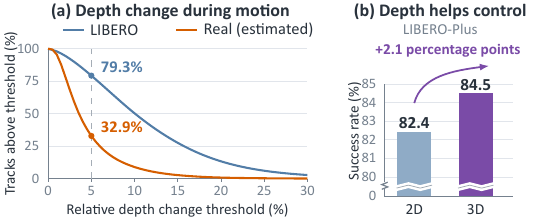}
\caption{\textbf{Depth complements image-plane motion.}
(a) Depth-change survival curves over 32-step point tracks whose maximum 2D displacement is at least 2\% of the image diagonal. Depth change is $\max_k |d_{t+k}-d_t|/d_t$. LIBERO uses simulator depth; Real uses estimated relative depth. (b) Retaining depth in motion supervision improves LIBERO-Plus success (\cref{tab:pending_controls}).}
\label{fig:depth_motivation}
\end{figure}

We therefore focus on \textbf{3D motion} as an auxiliary prediction target for WAM representation learning, motivated by two complementary properties. First, each trajectory connects a point in the current observation to its positions at successive future steps, revealing the coordinated motion of the robot and task-relevant objects (\cref{fig:teaser}(a)). Second, depth change complements image-plane displacement by describing motion toward or away from the camera. 
In our annotation census, among point tracks moving by at least 2\% of the image diagonal over 32 steps, 79.3\% in LIBERO and 32.9\% in Piper exhibit depth changes of at least 5\% of their initial depth (\cref{fig:depth_motivation}(a)). Depth variation therefore accompanies a substantial fraction of visible motion in both simulation and real-world annotations. Its value for policy learning is supported by our depth ablation: retaining the depth component improves LIBERO-Plus success from 82.4\% to 84.5\% (\cref{fig:depth_motivation}(b)).
These properties describe the temporal and spatial changes involved in manipulation.

\begin{figure*}[t]
\centering
\includegraphics[width=\textwidth]{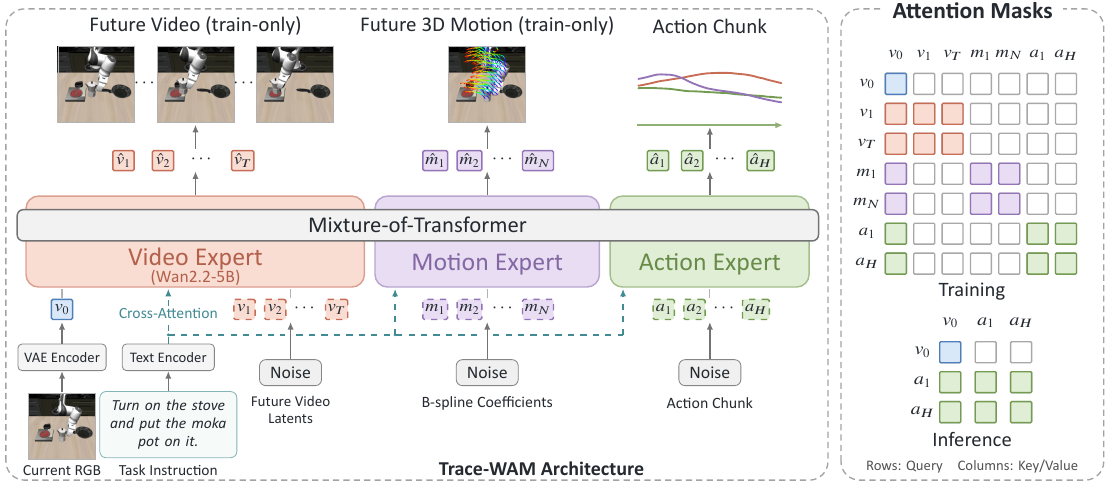}
\caption{\textbf{Overview of Trace-WAM.} Left: three Transformer experts jointly learn future video prediction, 3D motion prediction, and action generation. Right: colored cells allow attention among the illustrated video, motion, and action streams. At inference, future-video and motion prediction are removed, and action denoising uses cached visual keys and values.}
\label{fig:architecture}
\end{figure*}

Based on this formulation, we propose \textbf{Trace-WAM}, which jointly learns future video prediction, future 3D motion prediction, and action generation. We represent motion using trajectories anchored to a fixed grid in the current image, combining image-plane displacement with depth change relative to each point's initial position. Whereas 4D-WAM~\cite{yang20264d} transfers trajectory knowledge by aligning temporal feature changes and source-to-destination relations with those of a pretrained 4D tracking model, Trace-WAM directly predicts motion trajectories under an auxiliary objective. This objective adds explicit supervision for subsequent robot and object motion to video-action representation learning.
Concretely, Trace-WAM uses video, motion, and action experts with a structured attention mask (\cref{fig:architecture}). During training, the video, motion, and action objectives jointly optimize the model, using future 3D motion as an additional learning signal for the representations used in action generation. At inference, we retain the Fast-WAM~\cite{yuan2026fast} policy path: the video backbone processes current RGB observations once, and the action expert denoises an action chunk using cached visual features, language, and proprioception. As summarized in \cref{fig:teaser}(a), Trace-WAM requires no future-video or motion prediction at inference.

We evaluate Trace-WAM on RoboCasa-GR1, LIBERO, and LIBERO-Plus, covering both in-distribution control and out-of-distribution generalization. Trace-WAM achieves strong performance across these settings, with particularly clear gains over Fast-WAM when demonstrations are limited and under distribution shifts (\cref{fig:teaser}(b)). Experiments on a real robot further show higher success rates across three physical manipulation tasks. These results support 3D motion prediction as an effective auxiliary objective for WAM policy learning. Our contributions are:
\begin{itemize}[leftmargin=*,itemsep=1pt,topsep=2pt]
 \item We introduce Trace-WAM, which jointly learns future video prediction, future 3D motion prediction, and action generation within a three-branch architecture.
 \item Future 3D motion provides auxiliary supervision for video-action representation learning without requiring future video or motion prediction at inference.
 \item Experiments on three manipulation benchmarks demonstrate strong in-distribution performance and out-of-distribution generalization, while real-world evaluations show consistent gains across three physical tasks.
\end{itemize}

\section{Related Work}
\label{sec:related}

\subsection{World Action Models}
Early approaches formulate manipulation as visual planning, using generative models to propose future observations and learned controllers to translate the resulting visual plans into executable actions~\cite{du2023learning,black2024zero}. More recent World Action Models integrate video priors more directly into policy learning, either conditioning actions on predictive video features~\cite{hu2024video,liao2026genie} or jointly modeling future visual observations and action sequences~\cite{ma2026dit4dit,yan2026robust,li2026causal,bi2026motus,kim2026cosmos,ye2026world,team2026motubrain}.

Beyond RGB prediction, recent methods extend predictive supervision to semantic features, depth and 3D geometry, and motion, encouraging policy representations to capture object identity, spatial layout, and scene dynamics~\cite{yuan2026dreamwam,ma2026geosem,yan2026robust,wang2026st,li2026egowam,yan2608flex}. Complementary supervision of affordances, object-centric interaction configurations, and goal regions emphasizes which scene elements are involved in the task and where the robot should interact~\cite{zhang2026deva,zheng2026gift}. Trace-WAM follows this direction by using auxiliary future 3D motion prediction to supervise temporal point correspondence and depth change jointly with future RGB prediction and action generation.

\subsection{Motion for Policy Learning}
Motion representations connect action generation to the spatial and temporal structure of manipulation by describing how the robot and objects move throughout an interaction. Incorporating depth further captures displacement along the viewing direction, complementing image-plane motion with spatial information needed for reaching, grasping, and object transport. Existing approaches couple action learning with optical flow or 2D point-motion prediction, using motion features or predicted motion to guide control~\cite{chen2026flowwam,yang2026mt,wang2026mowam}, while others incorporate depth and 3D motion to model spatial scene dynamics for action generation~\cite{zhao2026rynnworld,wang2026lamp,lee2026mu_0}.
A complementary line of work uses motion supervision to improve policy representations, through distillation from pretrained tracking models~\cite{wang2026track4action,yang20264d} or auxiliary 3D motion prediction during human--robot co-training~\cite{li2026egowam}. Trace-WAM directly predicts future 3D point trajectories alongside RGB prediction and action generation, using numerical motion targets rather than tracker-feature alignment. The motion objective shapes video-action representations during training; deployed actions use current visual context, language, and proprioception without future-video or motion prediction.

\section{Method}
\label{sec:method}

Trace-WAM jointly learns future video prediction, future 3D motion prediction, and action generation. Future 3D motion provides auxiliary supervision for video-action representation learning, while the deployed policy generates actions from current observations. \Cref{fig:architecture} summarizes the Tri-MoT architecture and its attention masks.

\subsection{Problem Formulation}
\label{sec:formulation}
We consider language-conditioned robot manipulation. At each decision step, the policy receives current RGB observations $I_0$, a language instruction $\ell$, and the robot's proprioceptive state $s_0$. It generates an action chunk
\begin{equation}
 A\sim\pi_\theta(\cdot\mid I_0,\ell,s_0),
 \qquad A\in\mathbb{R}^{H_A\times D_A},
 \label{eq:policy}
\end{equation}
where $H_A$ is the action horizon and $D_A$ is the action dimension. The robot executes the first few actions before receiving a new observation and replanning.

We follow Fast-WAM~\cite{yuan2026fast} in training the policy with future video prediction while generating actions from current observations. Trace-WAM adds future 3D motion prediction as an auxiliary objective. Future RGB and motion labels are used only during training.

\subsection{3D Motion Representation}
\label{sec:trajectory}
\paragraph{Motion Targets.}
Following the grid-based motion representation of LaMP~\cite{wang2026lamp}, we place $Q$ anchors on a uniform grid covering the current main-camera image and associate each anchor with its positions over $H_M$ future steps. Our reference configuration uses a $20\times20$ grid ($Q=400$) and $H_M=32$, giving 33 positions per anchor when the current frame is included. Anchors follow a fixed row-major order. Although the policy can receive multiple camera views, the motion grid is defined only in the main-camera image.

For anchor $q$ at step $t$, let $p_{q,t}=(u_{q,t},v_{q,t},d_{q,t})$ contain image coordinates normalized to $[-1,1]$ and depth, measured in meters for simulator annotations and in an aligned relative scene scale for TraceForge annotations. Motion is displacement relative to its current-image position:
\begin{equation}
 \Delta p_{q,t}=p_{q,t}-p_{q,0},
 \qquad t=0,\ldots,H_M,
 \label{eq:motion}
\end{equation}
where $\Delta p_{q,0}=0$. The resulting target $\Delta P\in\mathbb{R}^{Q\times(H_M+1)\times3}$ combines image-plane displacement with depth change. Every displacement is measured from the current position, rather than accumulated from adjacent frames. All trajectories are expressed in a unified reference-camera frame before conversion to $(u,v,d)$, compensating for camera motion. This screen-aligned representation retains the spatial information provided by depth without depending on a particular robot embodiment, while its $(u,v)$ coordinates remain compatible with 2D trajectory formats.
For simulated data, motion labels are generated from simulator geometry, depth, and entity poses, whereas real-world videos are processed by the TraceForge pipeline from TraceGen~\cite{lee2026tracegen} to estimate depth, camera motion, and point tracks; the supplementary material describes the complete annotation and validation procedure.

\paragraph{B-Spline Representation.}
Following $\mu_0$~\cite{lee2026mu_0}, we represent each motion sequence with cubic B-spline control points. After scaling each displacement channel using training-set statistics, we fit $K$ control points to the valid positions, including the initial zero displacement:
\begin{equation}
 \widetilde{\Delta P}_q\approx B C_q,
 \qquad C_q\in\mathbb{R}^{K\times3},
 \label{eq:spline}
\end{equation}
where $\widetilde{\Delta P}_q$ is the scaled motion sequence and $B$ is a fixed spline basis matrix. The reference configuration uses $K=10$ control points per anchor. The fitted coefficients provide a compact target for the motion expert. Multiplying predicted coefficients by $B$ and restoring the channel scales recovers the displacement sequence. This representation preserves the correspondence between each current-image anchor and its future motion.

\subsection{Model Architecture}
\label{sec:experts}
\paragraph{Tri-MoT Architecture.}
Trace-WAM uses three Transformer experts with separate parameters and input/output projections. A frozen causal video VAE encodes current and future RGB as latent tokens. The video expert processes clean current tokens $V_0$ and, during training, noisy future tokens $V_f$. The action expert embeds each step of a noisy action chunk as a token. A trainable projection embeds the current state as proprioceptive tokens; a frozen text encoder supplies instruction embeddings via cross-attention.

\paragraph{Motion Expert.}
The motion expert operates on noisy B-spline coefficients. We arrange the $3K$ coefficients of each anchor as channels on the spatial grid, then embed them with a convolution whose kernel and stride are both $2\times2$. With the reference settings, this maps the $20\times20$ grid to 100 motion tokens. Each token contains four complete future trajectories, preserving their individual coefficients. This packs the motion horizon into token channels instead of creating a token for every future time step.

We encode each patch center with two-dimensional Fourier features and add their learned projection to its token embedding. Noise-time embeddings modulate the Transformer normalization and gated residual updates. At every layer, motion tokens attend to one another and to current video features, while the instruction and proprioceptive tokens provide task and state conditions. The output head predicts $4\times K\times3$ velocity components per token, unpacked into per-anchor coefficient velocities. 

\paragraph{Structured Attention.}
At each layer, the experts project their queries, keys, and values to a common attention dimension. The attention masks in \cref{fig:architecture} allow current video tokens to attend only to one another, while future-video, motion, and action tokens attend to their own group and to current video. Attention within future video and within the action chunk is bidirectional across time.

The video, motion, and action objectives are jointly optimized, using future 3D motion as an additional learning signal for the representations used in action generation. Because neither current video nor action tokens attend to motion or future video, the deployed policy does not depend on either prediction branch.

\subsection{Training Objective}
\label{sec:training}
\paragraph{Flow Matching.}
We jointly train the three experts with flow matching~\cite{lipman2022flow}. Let $x_r$ denote the clean target for branch $r\in\{V,M,A\}$: future video latents, motion coefficients, or an action chunk. For each branch, Gaussian noise $\epsilon_r$ and noise level $\sigma_r\in[0,1]$ define
\begin{equation}
 x_{r,\sigma_r}=(1-\sigma_r)x_r+\sigma_r\epsilon_r,
 \qquad v_r^*=\epsilon_r-x_r.
 \label{eq:flow_target}
\end{equation}
The expert predicts velocity $v_{\theta,r}$ from this noisy target, the noise level, and the conditions allowed by the attention mask. Its basic loss is
\begin{equation}
 \loss_r^{\mathrm{FM}}=
 \mathbb{E}\!\left[w_r(\sigma_r)
 \left\|v_{\theta,r}-v_r^*\right\|^2\right],
 \label{eq:flow_loss}
\end{equation}
where $w_r$ is the noise-level-dependent weight. We use this loss for $\loss_V$ and $\loss_A$, with motion errors balanced below. 

\paragraph{Motion Supervision.}
Static anchors can greatly outnumber moving anchors. For each sample, we compute mean coefficient-velocity errors $\ell_{\mathrm{mov}}$ and $\ell_{\mathrm{stat}}$ for moving and static anchors separately and combine them:
\begin{equation}
 \loss_M=\mathbb{E}\!\left[w_M(\sigma_M)
 \left(\alpha\ell_{\mathrm{mov}}+(1-\alpha)\ell_{\mathrm{stat}}\right)\right].
 \label{eq:motion_balance}
\end{equation}
Here, $\alpha$ controls the two groups' contributions. 

We additionally penalize residual motion at static anchors in the decoded trajectory space. From the predicted velocity, we estimate clean coefficients as $\hat C_0=C_{\sigma_M}-\sigma_M v_{\theta,M}$ and decode them with the spline basis and channel scales to obtain $\widehat{\Delta p}_{q,t}$. The static-motion loss is
\begin{equation}
 \loss_{\mathrm{zero}}=
 \mathbb{E}_{\mathrm{stat}}\!\left[
 \underset{q\in\mathcal S}{\operatorname{mean}}\;
 \underset{t\in\mathcal T_q}{\operatorname{mean}}
 \rho_\beta\!\left(\widehat{\Delta p}_{q,t}\right)\right],
 \label{eq:static_motion}
\end{equation}
where $\mathcal S$ and $\mathcal T_q$ index static anchors and future times, respectively, and $\rho_\beta$ is a channel-averaged Smooth-L1 penalty. It directly encourages stationary points to remain stationary after decoding.

\paragraph{Joint Training.}
The full objective combines video prediction, balanced motion prediction, static-motion regularization, and action prediction:
\begin{equation}
 \loss=\lambda_V\loss_V+\lambda_M\loss_M
       +\lambda_{\mathrm{zero}}\loss_{\mathrm{zero}}+\lambda_A\loss_A,
 \label{eq:total}
\end{equation}
where the $\lambda$ terms set the loss weights. Training begins with a warmup using the video and motion objectives, including the static-motion loss. We then add the action objective and jointly optimize all three experts. The VAE and text encoder remain frozen; the three experts and proprioceptive-token projection are trainable.


\section{Experiments}
\label{sec:experiments}


\begin{table}[t]
\caption{\textbf{In-distribution performance on RoboCasa-GR1.} Success rates (\%) in the few-shot and full-data settings. P.T. indicates additional robotic pretraining. DIAL is trained without human data; Best and second-best results are shown in bold and underlined, respectively.}
\label{tab:gr1}
\centering
\fontsize{9}{10.5}\selectfont
\setlength{\tabcolsep}{4pt}
\begin{tabular*}{\columnwidth}{@{\extracolsep{\fill}}lccc@{}}
\toprule
Method & P.T. & Few-shot & Full \\
\midrule
GR00T N1.6~\cite{bjorck2025gr00t} & $\checkmark$ & -- & 47.6 \\
StarVLA-OFT~\cite{community2026starvla} & $\times$ & -- & 48.8 \\
StarVLA-$\alpha$~\cite{ye2026starvla} & $\times$ & -- & 53.8 \\
ABot-M0~\cite{yang2026abot} & $\checkmark$ & -- & 58.3 \\
RLDX-1~\cite{kim2026rldx} & $\checkmark$ & -- & 58.7 \\
JoyAI-RA~\cite{zhang2026joyai} & $\checkmark$ & -- & 63.2 \\
DIAL~\cite{chen2026dial} & $\times$ & 58.3 & 70.2 \\
\midrule
Being-H0.7~\cite{luo2026being} & $\checkmark$ & -- & 49.2 \\
DiT4DiT~\cite{ma2026dit4dit} & $\checkmark$ & -- & 50.8 \\
LDA-1B~\cite{lyu2026lda} & $\checkmark$ & -- & 55.4 \\
LAWA w/ ego~\cite{li2026latent} & $\checkmark$ & \textbf{65.6} & \textbf{80.8} \\
LAWA w/o ego~\cite{li2026latent} & $\times$ & 59.7 & 76.3 \\
Fast-WAM~\cite{yuan2026fast} & $\times$ & 44.8 & 74.6 \\
Joint-WAM~\cite{yuan2026fast} & $\times$ & 54.8 & 78.3 \\
\midrule
\textbf{Trace-WAM (Ours)} & $\times$ & \underline{63.6} & \underline{80.3} \\
\bottomrule
\end{tabular*}
\par\smallskip
\begin{minipage}{\columnwidth}
\fontsize{8}{9.2}\selectfont
\end{minipage}
\end{table}

\begin{figure}[t]
\centering
\includegraphics[width=\columnwidth]{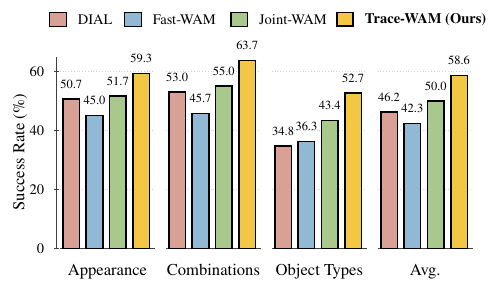}
\caption{\textbf{Few-shot OOD generalization on RoboCasa-GR1.}}
\label{fig:gr1_ood}
\end{figure}

\begin{table}[t]
\caption{\textbf{Quantitative comparison on LIBERO~\cite{liu2023libero}.} Success rates (\%) across four task suites.}
\label{tab:libero}
\centering
\fontsize{9}{10.5}\selectfont
\setlength{\tabcolsep}{1.5pt}
\begin{tabular*}{\columnwidth}{@{}l@{\hspace{4pt}}c@{\hspace{5pt}\extracolsep{\fill}}ccccc@{}}
\toprule
\multirow{2}{*}[-0.4ex]{Method} & \multirow{2}{*}[-0.4ex]{P.T.} & \multicolumn{4}{c}{Task Suite} & \multirow{2}{*}[-0.4ex]{Avg.} \\
\cmidrule(lr){3-6}
 & & Spatial & Object & Goal & Long & \\
\midrule
OpenVLA~\cite{kim2024openvla} & $\checkmark$ & 84.7 & 88.4 & 79.2 & 53.7 & 76.5 \\
OpenVLA-OFT~\cite{kim2025fine} & $\checkmark$ & 97.6 & 98.4 & 97.9 & 94.5 & 97.1 \\
UniVLA~\cite{bu2025univla} & $\checkmark$ & 96.5 & 96.8 & 95.6 & 92.0 & 95.2 \\
$\pi_0$~\cite{black2024pi_0} & $\checkmark$ & 96.8 & 98.8 & 95.8 & 85.2 & 94.2 \\
$\pi_0$-FAST~\cite{pertsch2025fast} & $\checkmark$ & 96.4 & 96.8 & 88.6 & 60.2 & 85.5 \\
$\pi_{0.5}$~\cite{intelligence2025pi_} & $\checkmark$ & 98.8 & 98.2 & 98 & 92.4 & 96.9 \\
\midrule
LingBot-VA~\cite{li2026causal} & $\checkmark$ & 98.5 & 99.6 & 97.2 & \textbf{98.5} & 98.5 \\
Motus~\cite{bi2026motus} & $\checkmark$ & 96.8 & \underline{99.8} & 96.6 & 97.6 & 97.7 \\
LaWAM~\cite{chen2026lawam} & $\checkmark$ & \underline{99} & 96 & 97.2 & 93.4 & 96.4 \\
IDM-WAM~\cite{yuan2026fast} & $\times$ & 98.6 & 99.4 & 98.4 & 96.4 & 98.2 \\
Fast-WAM~\cite{yuan2026fast} & $\times$ & 97.2 & \textbf{100} & 97 & 95.2 & 97.6 \\
Joint-WAM~\cite{yuan2026fast} & $\times$ & \textbf{99.6} & 99.4 & 98.2 & 96.8 & 98.5 \\
ImageWAM~\cite{zhang2026imagewam} & $\times$ & 97.2 & 99.2 & \underline{98.8} & \underline{98.4} & 98.4 \\
GIFT-WAM-Fast~\cite{zheng2026gift} & $\times$ & 97.8 & 99.6 & 98.6 & 94.8 & 97.7 \\
GeoSem-WAM~\cite{ma2026geosem} & $\times$ & \underline{99.0} & \textbf{100.0} & 98.2 & 97.0 & \underline{98.6} \\
\midrule
\textbf{Trace-WAM (Ours)} & $\times$ & 98.2 & \underline{99.8} & \textbf{99.2} & 97.6 & \textbf{98.7} \\
\bottomrule
\end{tabular*}
\par\smallskip
\begin{minipage}{\columnwidth}
\fontsize{8}{9.2}\selectfont
\end{minipage}
\end{table}

\begin{table*}[t]
\caption{\textbf{Zero-shot transfer from LIBERO to LIBERO-Plus~\cite{fei2025libero}.} Success rates (\%) across seven perturbation types.}
\label{tab:plus}
\centering
\fontsize{9}{10.5}\selectfont
\setlength{\tabcolsep}{2pt}
\begin{tabular*}{\textwidth}{@{}l@{\hspace{5pt}}c@{\hspace{5pt}\extracolsep{\fill}}cccccccc@{}}
\toprule
\multirow{2}{*}[-0.4ex]{Method} & \multirow{2}{*}[-0.4ex]{P.T.} & \multicolumn{7}{c}{Perturbation Type} & \multirow{2}{*}[-0.4ex]{Avg.} \\
\cmidrule(lr){3-9}
 & & Camera & Robot & Language & Light & Background & Noise & Layout & \\
\midrule
OpenVLA~\cite{kim2024openvla} & $\checkmark$ & 0.8 & 3.5 & 23.0 & 8.1 & 34.8 & 15.2 & 28.5 & 15.6 \\
OpenVLA-OFT~\cite{kim2025fine} & $\checkmark$ & 56.4 & 31.9 & 79.5 & 88.7 & \textbf{93.3} & 75.8 & 74.2 & 69.6 \\
UniVLA~\cite{bu2025univla} & $\checkmark$ & 1.8 & 46.2 & 69.6 & 69.0 & 81.0 & 21.2 & 31.9 & 42.9 \\
$\pi_0$~\cite{black2024pi_0} & $\checkmark$ & 13.8 & 6.0 & 58.8 & 85.0 & 81.4 & 79.0 & 68.8 & 53.6 \\
$\pi_0$-FAST~\cite{pertsch2025fast} & $\checkmark$ & 65.1 & 21.6 & 61.0 & 73.2 & 73.2 & 74.4 & 68.8 & 61.6 \\
\midrule
Fast-WAM~\cite{yuan2026fast} & $\times$ & 40 & 68.5 & 65.6 & 89.8 & 66.9 & 63.3 & 76.8 & 66.2 \\
Joint-WAM~\cite{yuan2026fast} & $\times$ & 59.4 & \underline{78.3} & \textbf{92.7} & 95.7 & 65.2 & 75.6 & \underline{84.5} & 78.7 \\
ImageWAM~\cite{zhang2026imagewam} & $\times$ & \textbf{80.8} & 50.3 & 91.4 & \textbf{98.1} & \underline{85.5} & \textbf{93.8} & 80.5 & \underline{83.1} \\
GIFT-WAM-Fast~\cite{zheng2026gift} & $\times$ & 37.4 & 75.3 & 79.5 & 94.5 & 70.3 & 78.7 & 78.8 & 72.6 \\
4D-WAM~\cite{yang20264d} & $\times$ & 45.15 & 64.26 & 90.63 & 94.29 & 57.71 & 69.08 & 79.21 & 71.01 \\
LAWA~\cite{li2026latent} & $\times$ & \underline{69.2} & 66.7 & 62.8 & 96.2 & 64.5 & 85.5 & 78.6 & 74.4 \\
\midrule
\textbf{Trace-WAM (Ours)} & $\times$ & 66 & \textbf{87.4} & 91.9 & \underline{97.2} & 76.1 & \underline{88.6} & \textbf{85.7} & \textbf{84.5} \\
\bottomrule
\end{tabular*}
\par\smallskip
\begin{minipage}{\textwidth}
\fontsize{8}{9.2}\selectfont
\end{minipage}

\end{table*}

\begin{table*}[t]
\caption{\textbf{Ablation Studies on LIBERO-Plus.} Success rates (\%) across seven perturbation types. (1) Motion supervision ablates moving/static balancing in $\loss_M$ and the static-motion penalty $\loss_{\mathrm{zero}}$. (2) Motion conditioning compares auxiliary supervision with Motion Inject and bidirectional video--motion attention. (3) Motion representation compares 2D/3D motion and raw-motion/B-spline targets. Best and second-best results are shown in bold and underlined, respectively.}
\label{tab:pending_controls}
\centering
\fontsize{9}{10.5}\selectfont
\setlength{\tabcolsep}{3pt}
\begin{tabular*}{\textwidth}{@{\extracolsep{\fill}}lcccccccc@{}}
\toprule
Method & Camera & Robot & Language & Light & Background & Noise & Layout & Avg. \\
\midrule
\rowcolor{blue!4}
\multicolumn{9}{c}{\textbf{\textit{1) Motion Supervision Ablation}}} \\
w/o Motion FM-split & 55.7 & 84.5 & 89.2 & 96.1 & 70.5 & 85.1 & 83.7 & 80.4 \\
w/o Static-Motion Penalty & 60.9 & \underline{86.9} & \underline{93.4} & \textbf{97.3} & \underline{75.3} & \underline{88.1} & 85 & \underline{83.6} \\
\midrule
\rowcolor{green!4}
\multicolumn{9}{c}{\textbf{\textit{2) Motion Conditioning Ablation}}} \\
Motion Inject & 61.7 & 82.8 & \underline{93.4} & 96.3 & 73.1 & 81 & 84.9 & 81.6 \\
Video--Motion Bid-Attention & 54.4 & 83.5 & \textbf{94} & 96.2 & 68.3 & 82.3 & 84.8 & 80.3 \\
\midrule
\rowcolor{yellow!8}
\multicolumn{9}{c}{\textbf{\textit{3) Motion Representation Ablation}}} \\
w/o Depth (2D Motion) & \underline{65.1} & 86.3 & 91.9 & 95.4 & 70.3 & 82.6 & \textbf{85.8} & 82.4 \\
w/o B-spline (Raw Motion) & 64.2 & 85.5 & 91.3 & \underline{97.2} & 75.1 & 86.8 & 85.4 & 83.4 \\
\midrule
\textbf{Trace-WAM (Ours)} & \textbf{66} & \textbf{87.4} & 91.9 & \underline{97.2} & \textbf{76.1} & \textbf{88.6} & \underline{85.7} & \textbf{84.5} \\
\bottomrule
\end{tabular*}
\par\smallskip
\begin{minipage}{\textwidth}
\fontsize{8}{9.2}\selectfont
\end{minipage}
\end{table*}

\newcommand{\MotionPredictionTable}{%
\begin{table}[t]
\caption{\textbf{Motion Prediction Accuracy on LIBERO-Plus.} We aggregate 140 rollout episodes at each horizon; lower is better.}
\label{tab:motion_prediction}
\centering
\fontsize{9}{10.5}\selectfont
\setlength{\tabcolsep}{4pt}
\begin{tabular*}{\columnwidth}{@{\extracolsep{\fill}}lccc@{}}
\toprule
Motion Representation & 8 Steps & 16 Steps & 32 Steps \\
\midrule
\multicolumn{4}{c}{\textit{Image-Plane ADE (pixels, $\downarrow$)}} \\
Raw Motion (Direct32) & 0.92 & 1.25 & 1.87 \\
\textbf{B-spline Motion} & \textbf{0.67} & \textbf{0.95} & \textbf{1.43} \\
\midrule
\multicolumn{4}{c}{\textit{Depth MAE (mm, $\downarrow$)}} \\
Raw Motion (Direct32) & 1.634 & 2.297 & 3.427 \\
\textbf{B-spline Motion} & \textbf{1.277} & \textbf{1.808} & \textbf{2.768} \\
\bottomrule
\end{tabular*}
\end{table}%
}

\subsection{Experimental Setup}

\paragraph{Benchmarks.}
We evaluate Trace-WAM on RoboCasa-GR1~\cite{nasiriany2024robocasa}, LIBERO~\cite{liu2023libero}, and LIBERO-Plus~\cite{fei2025libero} using task success rate. RoboCasa-GR1 contains 24 tabletop rearrangement and articulated-object manipulation tasks. We evaluate both full-data (24,000 trajectories, 1,000 per task) and few-shot settings (10\% subset), and test the few-shot policy on Unseen Appearance, Unseen Combinations, and Unseen Object Types following DIAL~\cite{chen2026dial}. LIBERO comprises four task suites: Spatial, Object, Goal, and Long. All models train solely on original LIBERO demonstrations and transfer zero-shot to LIBERO-Plus under seven perturbations: camera, robot initial state, language, lighting, background, sensor noise, and layout. No augmented LIBERO-Plus data are used for fine-tuning.

\paragraph{Implementation Details.}
We build on the Fast-WAM~\cite{yuan2026fast} codebase and training settings, using pretrained Wan2.2-5B~\cite{wan2025wan} as the video backbone. The action and motion experts each follow the video Transformer's architecture with a hidden dimension of 1,024. Both action and motion horizons are set to 32, and videos are temporally downsampled by $4\times$ to nine frames per chunk. For LIBERO and RoboCasa-GR1 few-shot, we train for 26,000 steps on 16 NVIDIA H20 GPUs with a global batch size of 128. For RoboCasa-GR1 full-data, we train for 50,000 steps on 32 NVIDIA H20 GPUs with a global batch size of 1,024. Each schedule starts with 2,000 steps of video--motion experts warmup, followed by joint training with the action expert for the remaining steps.


\subsection{Main Results}

\paragraph{Performance on RoboCasa.}
As shown in Table~\ref{tab:gr1}, Trace-WAM achieves success rates of 63.6\% and 80.3\% in the few-shot and full-data settings, outperforming Fast-WAM and Joint-WAM in both. The gains are especially pronounced with limited demonstrations, reaching 18.8 and 8.8 percentage points over the two baselines, respectively. Figure~\ref{fig:gr1_ood} further shows consistent generalization across unseen appearances, object combinations, and object types. Trace-WAM leads all three comparison methods in every setting and reaches 58.6\% average success, exceeding DIAL, Fast-WAM, and Joint-WAM by 12.4, 16.3, and 8.6 percentage points, respectively. Together, these results show that future 3D motion supervision supports both in-distribution performance and OOD generalization.

\paragraph{Performance on LIBERO.}
Trace-WAM achieves an average success rate of 98.7\% across the four LIBERO suites (Table~\ref{tab:libero}), compared with 97.6\% for Fast-WAM and 98.5\% for Joint-WAM. This further demonstrates strong in-distribution performance across diverse manipulation tasks.

\paragraph{Zero-shot Robustness on LIBERO-Plus.}
Without further adaptation, Trace-WAM reaches 84.5\% success on LIBERO-Plus (Table~\ref{tab:plus}), exceeding Fast-WAM and Joint-WAM by 18.3 and 5.8 percentage points, respectively. It improves over Fast-WAM across all seven perturbation types. In particular, success rates under Robot and Background perturbations reach 87.4\% and 76.1\%, improving over Fast-WAM by 18.9 and 9.2 points, respectively. These gains support the value of motion supervision for learning interaction-relevant representations: temporal correspondence and depth change describe how the robot and objects move during manipulation, encouraging features that remain useful under altered robot initial states and task-irrelevant background changes.

\subsection{Ablation Study and Analysis}
\label{sec:ablation}

\paragraph{Motion Supervision Ablation.}
Table~\ref{tab:pending_controls} shows the contributions of the two motion-supervision components. The \emph{w/o Motion FM-split} variant removes the separate balancing of moving and static anchors in the motion flow-matching loss, achieving 80.4\% average success. The \emph{w/o Static-Motion Penalty} variant achieves 83.6\%. Trace-WAM reaches 84.5\%, improving over these variants by 4.1 and 0.9 percentage points, respectively. These gains highlight the importance of balancing supervision from moving and static anchors, with static-motion regularization providing a further improvement in task success.

\paragraph{Motion Conditioning Ablation.}
The \emph{Motion Inject} variant injects future 3D motion latents from each motion-expert layer into the corresponding action-expert layer, analogous to the layerwise video--action connection in Joint-WAM, and achieves 81.6\% average success. The \emph{Video--Motion Bid-Attention} variant instead allows the video and motion streams to attend to each other at every layer and achieves 80.3\%. Trace-WAM keeps future 3D motion as a jointly optimized auxiliary objective and reaches 84.5\%, improving over these variants by 2.9 and 4.2 percentage points, respectively. These results favor auxiliary 3D motion supervision within video-action representation learning, without bidirectional video--motion coupling or action conditioning on future motion latents.

\paragraph{Motion Representation Ablation.}
Trace-WAM outperforms \emph{w/o Depth (2D Motion)} and \emph{w/o B-spline (Raw Motion)}, which achieve 82.4\% and 83.4\% average success, respectively. Adding depth yields a 2.1-point gain, with improvements of 5.8 and 6.0 points under Background and Noise perturbations. This supports the value of 3D motion supervision beyond image-plane displacement for OOD generalization. B-spline targets improve success by 1.1 points over direct per-step displacements, supporting compact motion prediction more effective for representation learning.

\MotionPredictionTable

\begin{figure*}[t]
\centering
\includegraphics[width=\textwidth]{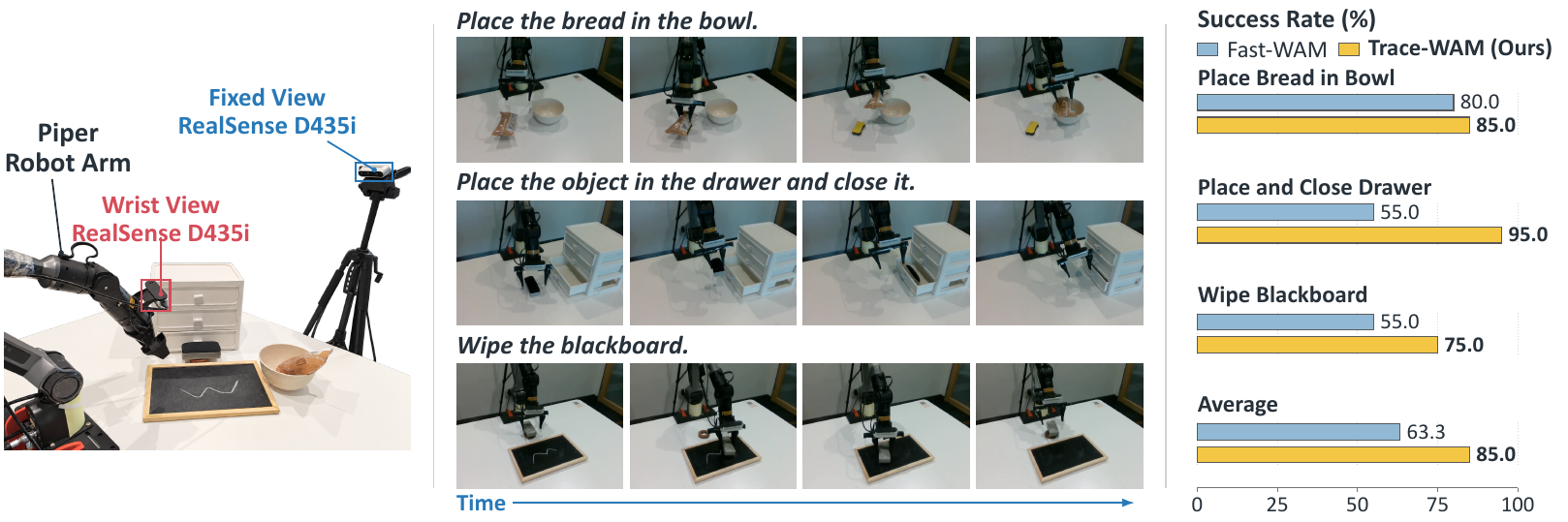}
\caption{\textbf{Real-world Experiments.} Left: a Piper arm with fixed-view and wrist-view RealSense D435i cameras. Center: four observed frames from a successful Trace-WAM execution of each task. Right: task success rates over 20 trials per task.}
\label{fig:piper_realworld}
\end{figure*}

\begin{figure}[t]
\centering
\includegraphics[width=\columnwidth]{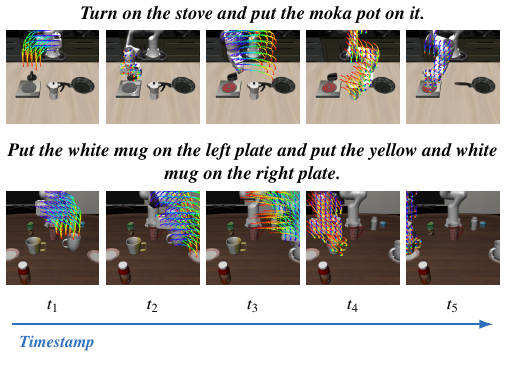}
\caption{\textbf{Qualitative future 3D motion predictions on LIBERO-Plus.} The overlays visualize B-spline motion predictions over a 32-step horizon.}
\label{fig:motion_prediction_vis}
\end{figure}

\paragraph{Motion Prediction Accuracy.}
We compare predicted motion with the motion realized in the simulator after executing the policy actions on LIBERO-Plus. Table~\ref{tab:motion_prediction} reports image-plane average displacement error (ADE) and depth mean absolute error (MAE) over all valid query--time pairs at 8-, 16-, and 32-step horizons. Both representations closely match the executed trajectories. Even at 32 steps, Raw Motion attains an image-plane ADE of 1.87 pixels and a depth MAE of 3.427~mm, while B-spline Motion reaches 1.43 pixels and 2.768~mm. The B-spline representation consistently lowers both errors across all three horizons, showing that its compact trajectory parameterization more accurately captures the realized 3D motion. Figure~\ref{fig:motion_prediction_vis} further visualizes B-spline predictions at five ordered replanning steps for two tasks. The predicted trajectories change with the ongoing robot--object interaction as each rollout progresses. Together with its higher task success in Table~\ref{tab:pending_controls}, these results support B-spline motion as the more effective representation for future 3D motion prediction and downstream action generation.

\subsection{Real-world Experiments}
\label{sec:real_world}

\paragraph{Setup.}
The experiments are conducted on the Piper platform with a gripper and two Intel RealSense D435i cameras providing fixed and wrist views (Figure~\ref{fig:piper_realworld}). 
We evaluate three tasks: \emph{Place Bread} requires placing bread in a bowl; \emph{Place and Close Drawer} requires placing an object in the drawer and closing it; and \emph{Wipe Blackboard} requires erasing the chalk markings.
Each task uses 50 demonstrations, and each method is evaluated over 20 trials per task. 

\paragraph{Results.}
Trace-WAM achieves 85.0\% average success across the three physical tasks, compared with 63.3\% for Fast-WAM (Figure~\ref{fig:piper_realworld}). The largest gain occurs on Place and Close Drawer, where success rises from 55.0\% to 95.0\%. Wipe Blackboard improves from 55.0\% to 75.0\%, and Place Bread in Bowl from 80.0\% to 85.0\%. These results extend the benefits of auxiliary 3D motion supervision to physical manipulation, including sequential placement and closing as well as sustained surface contact.

\section{Conclusion}
\label{sec:conclusion}

We presented Trace-WAM, which uses auxiliary future 3D motion prediction to focus video--action representation learning on the scene dynamics most relevant to manipulation. The video, motion, and action objectives are jointly optimized during training, while the deployed policy generates actions from current observations, language, and proprioception without running future-video or motion prediction. Across LIBERO, zero-shot LIBERO-Plus, and RoboCasa-GR1, Trace-WAM achieves strong in-distribution and out-of-distribution performance, with clear gains in the few-shot and OOD settings. Real-world experiments on a Piper robot further demonstrate improved success across three manipulation tasks. These results establish future 3D motion prediction as an effective training signal for learning robust action representations in WAMs.

{\small
\bibliographystyle{ieeenat_fullname}
\bibliography{main}
}
\end{document}